\documentclass[runningheads]{llncs}
\usepackage{graphicx}
\usepackage{enumerate}
\begin{document}
\title{Deep Learning Approach for Enhanced Cyber Threat Indicators in Twitter Stream}
\titlerunning{Deep Learning based Cyber Threat Indicators in Twitter}
%
%

\author{Simran K\inst{1}\and Prathiksha Balakrishna\inst{2} \and
Vinayakumar R\inst{3, 1}\and
Soman KP\inst{1}}
\authorrunning{Simran et al.}
%
\institute{Center for Computational Engineering and Networking, Amrita School Of Engineering, Amrita
vishwa vidyapeetham, Coimbatore, India. \email{simiketha19@gmail.com} \and Graduate School, Computer Science Department,\\ Texas State University. \email{prathi.93april8@gmail.com}\and Division of Biomedical Informatics, Cincinnati Children's Hospital Medical Centre, Cincinnati, OH, United States.\\ \email{Vinayakumar.Ravi@cchmc.org, vinayakumarr77@gmail.com}}

\maketitle             
\begin{abstract}
In recent days, the amount of Cyber Security text data shared via social media resources mainly Twitter has increased. An accurate analysis of this data can help to develop cyber threat situational awareness framework for a cyber threat. This work proposes a deep learning based approach for tweet data analysis. To convert the tweets into numerical representations, various text representations are employed. These features are feed into deep learning architecture for optimal feature extraction as well as classification. Various hyperparameter tuning approaches are used for identifying optimal text representation method as well as optimal network parameters and network structures for deep learning models. For comparative analysis, the classical text representation method with classical machine learning algorithm is employed. From the detailed analysis of experiments, we found that the deep learning architecture with advanced text representation methods performed better than the classical text representation and classical machine learning algorithms. The primary reason for this is that the advanced text representation methods have the capability to learn sequential properties which exist among the textual data and deep learning architectures learns the optimal features along with decreasing the feature size.

\end{abstract}
\section{Introduction}

As social media is an interactive platform where individuals share thoughts, information, professional interests and different types of expression via virtual communities and systems, it introduces a rich and timely source of information on events occurring everywhere throughout the world \cite{13}. Social media giants like Facebook, Twitter, WhatsApp, etc enable a lot of applications like recognizing the area of missing people during catastrophic events or earthquake detection. 
Past work on event extraction has depended on a large amount of labeled information or taken an open-domain approach in which general events are extracted without a particular core interest. Information analyst is often interested in tracking a very specific type of event, for example, data breaches or account hijacking and probably won't have time or expertise to build an information extraction framework from scratch in response to emerging incidents. To address this challenge we introduce a deep learning approach for rapid training cyber threat indicators for the Twitter stream.

Open-Source Intelligence (OSINT) provides a vital source of information and has proven to be an important asset for Cyber Threat Intelligence (CTI). One of OSINT rich sources is Twitter. Twitter's popularity in the society provides an environment for defensive and offensive Cyber Security practitioners to debate, and promote timely indicators of different type of cyber events such as attacks, malware, vulnerabilities, etc. Various initial reports of recent major cyber events like the exposure of multiple \textquotedblleft 0-day\textquotedblright    Microsoft Windows vulnerabilities, exposure of ransomware campaigns \cite{1} and user reports on DDoS attacks \cite{2} exhibits the value of Twitter data to CTI analysts. 

Multiple frameworks for detecting as well as analysing the treat indicators in Twitter stream have come from the research on Twitter-based OSINT collection. 
However, most of these proposals have a high false-positive rate in detecting the relevant tweets as they are using heavily manual heuristics like keyword lists that are relevant to Cyber Security are used to detect and filter tweets. Furthermore, potentially valuable information in tweets is getting neglected by the emergence of new terminology and flexible typography. In recent days, the applications of deep learning with natural language processing methods leveraged in various Cyber Security tasks \cite{13,14,15,16,17,extra1,extra2,extra3}. These methods have obtained good performance and most importantly, these methods performed well compared to the classical machine learning classifiers.   

The major contribution of this proposed work are given below:
\begin{enumerate}
    \item This work proposes a deep learning based framework for cyber threat indicators in the Twitter stream. The framework is highly scalable on using commodity hardware.
    \item To identify a proper tweet representation, various state-of-the-art text representation exists in the domain of natural language processing (NLP) are leveraged for cyber threat indicators in Twitter Stream.
    \item To identify an optimal machine learning approach, we have carried out a comprehensive and in-depth study of the application of classical machine learning and deep learning theory in the context of cyber threat indicators in Twitter stream.
    \item In particular, we discuss several parameterization options for classical machine learning, deep learning, and tweet representation and we present a large variety of benchmarks which have been used to either experimentally validate our choices or to help us to take the adequate decision.
\end{enumerate}

The remaining of this paper is arranged in the following order:
Section 2 documents a survey of the related literature, followed by background related to NLP and deep learning concepts in Section 3. Section 4 provides a description of Cyber Security related tweets data set used in this work. Section 5 describes the details of the proposed architecture. Section 6 reports the experiments and observations made by the proposed architecture. Section 7 concludes the paper as well as tells the remakes on future work of research.

\section{Literature Survey}

Classification and detection of CTI extraction from Twitter are less investigated compared to the other area, for example, crime prevention \cite{3}, identification of cyber-bullies \cite{4}, and disaster response \cite{5}. To distinguish three sorts of threats and events ie., account hijacking, data breaches, and Distributed Denial of Service (DDoS) attacks, Khandpur et al. \cite{6} proposed an architecture to separate cyber threat as well as security information from the Tweets. Target domain generation,  event extraction, and dynamically typed query expansion are the three major segments of this framework. This methodology is powerful as it abuses syntactic, semantic analysis and dependency tree graph yet it requires the persistent tracking of features for each type of threat. It likewise requests a high computational overhead to produce as well as keep the focus on corpus space of tweet content for query extension. Also, this architecture can't flawlessly stretch out to more classifications of threats and events.

Furthermore, categorizing Cyber Security events from tweets was proposed by Le Sceller et al. \cite{7}. The detection of events uses the taxonomy of Cyber Security and a set of keywords that describes the event type. Expanding of the set of seed keywords are performed by not only identifying but also attaching new words with comparative meaning with regards to word embeddings utilizing a physically indicated edge in the cosine similarity distance between word vectors. Term frequency - inverse document frequency (TF-IDF) method which produced events as groups of tweets was used in this framework. Inadvertent biasing effects of the initial seed keywords caused this algorithm to give a high false-positive rate.  

Security Vulnerability Concept Extractor (SVCE) was utilized to process tweets in the structure proposed in \cite{8}. SVSE is trained on a  data set containing reports of the national vulnerability database to recognize as well as label the terms and ideas identified with CTI, for example, affected software, consequences of the attack and so forth. To additionally improve the extracted information, the ideas and substances extracted by SVCE are examined dependent on outside freely accessible semantic learning bases such as DBPedia. The client needs to specify a target framework profile included data about installed software or hardware, as this system is produced for client-based applications. As per the information provided, an ontology is produced and utilized alongside SWRL rules to address as well as organize time-delicate CTI entries. Later conversion from separated and labeled CTIs to RDF triple proclamations is finished. The ready alert system can reason over the information as RDF connected information portrayal is put away in the knowledge base. This framework is incapable of distinguishing novel threat types and indicators.

In \cite{9}, ontology-based technique and Named Entity Recognition (NER) were utilized to classify tweets as related events or not related events. This framework performs topic identification by means of cross-referencing NER results with other external knowledge bases, for example, DBPedia utilizing Wikipedia’s Current Event Portal just as human info gathered using Amazon Mechanical Turk, produced an annotated data set of tweet event type and CTI. Different machine learning algorithms, for example, naive bayes, support vector machines (SVM) and deep learning architectures such as long short term memory (LSTM), recurrent neural network (RNN) used this annotated  data set and the best outcome was delivered by LSTM architecture with word embedding. They additionally show that particular classifications of NER are useful in classifying the classes as well as event type, though the nonexclusive class of NER is useful in binary classification. Pagerank algorithm was used in this work for topic recognizable proof.

\cite{10} proposed a framework which recognizes influential user or community of people to prioritize CTI information. This was finished utilizing a scoring strategy that is scores were given to the user and community who produced CTI-related tweets. This work has four segments. For gathering information from the Twitter platform, a social media connector is referred as the principal segment. The second segment is a module for recognizing and stretching out the rundown of specialists to discover developing themes. The third segment comprises of weight contribution and fitness calculation. Lastly, to recognize emerging threats the author proposed a topic detection algorithm. Anyhow, threat indicators are not adequately referred by the specialists in this work.

The framework proposed in \cite{11} is a weekly supervised learning approach that trains a model for extracting new classes of Cyber Security events. This framework does extraction by seeding a little amount of positive event tests over a fundamentally amount of unlabeled data. The target to learn in this work is done by regularizing the label distribution over the unlabeled distribution. This work is vigorously reliant on historical seed and neglects to give the details of coordinating named entities into an event category.

\section{Background}

\subsection{Text representation}

To represent the tweet into numeric form, we used various text representations in this works. The basic idea behind these text representations is discussed below.

\subsubsection{Bag-of-Words (TDM, TF-IDF):}

Bag-of-words is basically a collection of words. So the texts (tweets) are represented as a bag of its words. Every unique word passed as an input will have a position in this bag (vector). The vector records the frequency of the words in the tweets. Term document matrix (TDM) and Term frequency-inverse document frequency (TF-IDF) are features extracted from the documents. They are the measures used to understand the similarities between the tweets. TDM will have each corpus word as rows and document (tweet) as columns. The matrix represents the frequencies of the words occurring in that particular tweet. The most used words are highlighted because of their high frequency. TF-IDF tells how frequently a word occurs in a specific record contrasted with the whole corpus. The uncommon words are featured to demonstrate their relative significance.

\subsubsection{N-gram:}

N-gram is a contiguous order of n items from a given sample of content (tweets). N-gram with $N = 1$ is known as a unigram and it takes one word/character at once. $N = 2$ and $N=3$ are called bigram and trigram respectively and will take two and three words/characters at a time. If n words/characters are to be taken at once then N will be equal to n.

\subsubsection{Keras Embedding:}

Word Embedding basically converts words into a dense vector of real numbers in such a way that sequence and word similarities are additionally safeguarded. Keras offers an Embedding layer which is initialized with random weights. It will learn embedding of all the words in the training set but the input word should be represented by a unique integer. Keras is an open-source neural network library which contains various executions of generally utilized neural network building blocks. It also supports convolutional, recurrent neural networks and other common utility layers like pooling, batch normalization, and dropout.

\subsubsection{fastText:}

fastText chips away at character n-gram level instead of just word level (word2vec) and it is better for morphologically rich dialects. To convert words into vectors it utilizes skip-gram and subword model. Given the present word, skip-gram model predicts the surrounding words. In the event that window size is 2, at that point we see just 5 vectors at once. The subword model will see the internal structure of the words. In this model n-grams per word are extracted. For example, ‘her’ will have distinctive vector and n-gram ‘her’ from the word ‘where’ will have a different vector.

\subsection{Deep learning}

To understand which deep learning approach works for enhanced cyber threat indicators in the Twitter stream, we used various deep learning architectures. The basic idea behind different deep learning approaches is given below.

\subsubsection{Deep Neural Network:}

A deep neural network (DNN) is a neural network with multiple layers which makes it somewhat mind-boggling.  DNN contains one input layer, at least one hidden layer, and one output layer. Each hidden layer contains a rectified linear unit (ReLU). ReLU is an activation function which characterized the positive piece of its argument. It has less vanishing gradient problems and computationally efficient. Hidden layer is also called a fully connected layer since every neuron in one layer is associated with every neuron in the following layer. 

\subsubsection{Convolutional Neural Network:}

A convolutional neural network (CNN) otherwise called ConvNet is a deep neural network which is based on shared-loads architecture. It lessens the number of free parameters enabling the network to be deeper with fewer parameters. Generally, CNN architecture contains convolution, pooling, and fully connected layers. The convolution operation is performed using a number of filters which slide through the input and learns the features of the input data. Pooling layer is used to decrease the size of the feature matrix. The pooling can be min, max, or average. At the end of the CNN, there will be at least one fully connected layer where all the neurons are connected to all the neurons of its previous layer. Also in between these layers batchnomralization and dropout can be used. Batch normalization layer allows the network to learn by itself a little bit more independently of other layers and in turn reduces overfitting as it has slight regularization effects. Dropout is a regularization technique in which some neurons are randomly ignored during training the model. This method is treated like a layer and makes neural networks with different architectures to train in parallel.

\subsubsection{Recurrent Structures (RNN, LSTM, GRU):}

A recurrent neural network (RNN) is a recurrent structure where associations between nodes form a directed graph along a sequence. This enables RNN to display temporal dynamic behavior for a time sequence that is applied to natural language processing (NLP). RNNs can utilize their internal state to process arrangements of inputs yet can do it for just a short amount of time i.e., they can not remember long term data.

Long short-term memory (LSTM) network is another recurrent structure that contains a cell, and three gates namely, input, output, and forget gate. A cell recalls esteems over discretionary time intervals and the three gates direct the stream of data in and out of the cell. This makes LSTM remember long term information. LSTMs were created to manage the vanishing and exploding gradient problems that can be experienced when training conventional RNNs.

Gated recurrent unit (GRU) is an enhanced version of standard RNN and is also considered as a minor variation from LSTM. To tackle the disappearing gradient problem of a standard RNN, GRU utilizes update gate and reset gate. These two vectors choose what information ought to be passed to the output. They are exceptional in light of the fact that they can be trained to keep information from a long prior time, without washing it through time or evacuate information which is superfluous to the expectation.

\section{Description of the Data set}

The data set for data analysis of tweets from Twitter social media resource is provided by \cite{12}. The authors used a stream listener to listen to the streaming of tweets from Twitter. They selected a set of keywords in order to filter as well as narrow down the results of the stream listener. The words like \textquotedblleft 0day\textquotedblright \hspace{0.3mm} and \textquotedblleft vulnerability\textquotedblright \hspace{0.3mm} were selected for applicability to CTI. For producing more targeted filters, words related to a particular type of threat were selected. Preprocessing of the data set is also performed in \cite{12}. The detailed statistics are tabulated in Table \ref{Tab:1} and \ref{Tab:2}.

\begin{table}[!h]
\renewcommand{\arraystretch}{1.3}
\centering
\caption{Binary class Twitter data samples.}
\label{Tab:1}
\scalebox{1.1}{\begin{tabular}{|l|l|l|}
\hline
 \textbf{Data set}                    & \textbf{Relevant} & \textbf{Irrelevant} \\ \hline
\textbf{Train}            & 11,781             & 5,313                   \\ \hline
\textbf{Test}            & 2,989             & 1,285               \\ \hline

\end{tabular}}
\end{table}

\begin{table}[!h]
\renewcommand{\arraystretch}{1.3}
\centering
\caption{Multiclass Twitter data samples.}
\label{Tab:2}
\scalebox{1.1}{\begin{tabular}{|l|c|c|}
\hline
 \textbf{Category}                    & \textbf{Train data set} & \textbf{Test data set} \\ \hline
\textbf{Vulnerability}            & 5,926             & 1,428                   \\ \hline
\textbf{Ransomware}            & 2,549             & 654               \\ \hline
\textbf{DDoS}            & 1,776             & 469               \\ \hline
\textbf{Data leak}            & 106              & 30                    \\ \hline
\textbf{General}            & 5,588             & 1,410               \\ \hline
\textbf{Day}            & 585              & 145                  \\ \hline
\textbf{Botnet}            & 564              & 138                \\ \hline
\end{tabular}}
\end{table}

\section{Proposed Architecture}

\begin{figure}[!htbp]
  \centering
  \includegraphics[]{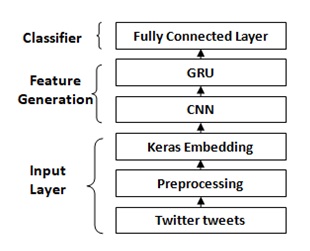}
  \caption{Proposed Architecture.}
  \label{Fig:1}
\end{figure}

The proposed architecture is shown in Figure \ref{Fig:1}. The preprocessed tweets are sent to Keras embedding layer where the words are converted into dense vectors. These numerical features are passed into CNN and then to GRU layer for feature generation. Finally, the output from GRU is sent to a fully connected layer for classification.

\section{Experiments, Results and Observations}

Scikit-learn\footnote{https://scikit-learn.org/} and TensorFlow\footnote{https://www.tensorflow.org/} with Keras\footnote{https://keras.io/} framework were utilized to implement classical machine learning algorithms and deep learning architectures respectively. All the models are trained on GPU enabled TensorFlow. Various statistical measures are utilized in order to evaluate the performance of the proposed framework.

\begin{table}[!h]
\renewcommand{\arraystretch}{1.3}
\centering
\caption{Average performance metrics.}
\label{Tab:5}
\begin{tabular}{|l|l|l|l|l|} \hline 
\textbf{Model}                    & \textbf{Accuracy (\%)} & \textbf{Precision (\%)} & \textbf{ Recall (\%)} & \textbf{F1-Score (\%)} \\ \hline

\multicolumn{5}{|c|}{\textbf{Binary class classification}}   \\ \hline
SVM-TDM                  & 81.9    & 68.8     & 72.8  & 70.7    \\ \hline
SVM-TF-IDF               & 82.2    & 69.2     & 73.6  & 71.3    \\ \hline
DNN-3gram                & 82.9    & 73.5     & 67.6  & 70.4    \\ \hline 
CNN-Keras word embedding \cite{12} & 83.6    & 71.4     & 75.9  & 73.6    \\  \hline
RNN-Keras word embedding        & 83.1    & 71.7     & 72.1  & 71.9    \\ \hline
LSTM-Keras word embedding                     & 84.3    & 70.1     & 83.1  & 76.0    \\  \hline
GRU-Keras word embedding                      & 84.7    & 73.9     & 76.0  & 74.9    \\ \hline
\textbf{CNN-GRU-Keras word embedding}                 &\textbf{85.8}    &\textbf{ 73.7 }    & \textbf{82.3}  &\textbf{77.8 }   \\ \hline
fastText                 & 84.4    & 74.6     & 73.2  & 73.9
\\ \hline

\multicolumn{5}{|c|}{\textbf{Multiclass classification}}   \\ \hline

SVM-TDM                  & 86.2    & 86.2     & 86.2 & 86.2    \\ \hline
SVM-TF-IDF               & 86.3    & 86.4     & 86.3  & 86.3    \\ \hline
DNN-3gram                &  86.9   &  87.0    & 86.9 &  86.9   \\ \hline 
CNN-Keras word embedding \cite{12} & 87.5   & 87.8     & 87.5 & 87.6    \\  \hline
RNN-Keras word embedding                      & 87.0    & 87.1     & 87.0  & 87.0    \\ \hline
LSTM-Keras word embedding                     & 88.0    & 88.1     & 88.0  & 88.0    \\  \hline
GRU-Keras word embedding                      & 88.4    & 88.8     & 88.4  & 88.5    \\ \hline
\textbf{CNN-GRU-Keras word embedding}                  & \textbf{89.3}    & \textbf{90.3}    & \textbf{89.3}  & \textbf{89.3}   \\ \hline
fastText                 & 87.9    & 88.0     & 87.9  & 87.9
\\ \hline

\end{tabular}
\end{table}

Preprocessing steps given in the proposed architecture section is followed for the data set to convert the unstructured format into a structured format. In this work, various text representation methods such as TDM, TF-IDF, 3-gram, and embedding are employed. SVM is implemented along with TDM and TF-IDF. SVM uses rbf kernel and c value of 100. Scikit-learn default parameters of TDM and TF-IDF are used. As the tweet length is not huge and there are a lot of important keywords used in tweets that might be the reason why TF-IDF has performed better than TDM. We followed n-gram representation specifically 3-gram is employed and we constructed a feature vector whose length will very huge. So in order to decrease the dimension be employed featurization technique to decrease the length of the sequence. This 1,000 length vector is passed into a deep neural network (DNN). DNN contains three layers, the first layer contains 1,024 neurons, the second layer contains 512 neurons and the third layer contains 128 neurons. In a sequential model, initially random weights are given to the model and these random values will be updated based on the loss of the function while backpropagation. When Keras embedding is employed along with the deep learning model, updation of weight will take place upto the embedding layer during backpropogation and not just stop at the deep neural layers. Since the  data set is not huge, word embedding like word2vec is not followed in this work. Various deep learning classifiers like CNN, RNN, LSTM, GRU, CNN-GRU are used along with Keras word embedding in order to find the best deep learning model. Embedding size of 128, batch-size of 32, learning rate of 0.01, 128 hidden units, and Adam optimizer are hyperparameter value used by RNN, LSTM, GRU, CNN, and CNN-GRU classifiers. The output layer consists of 1 neuron in binary classification and 7 neurons for multiclass classification. In CNN, the number of filters used is 64 and the filter length is 3. In CNN-GRU as well as GRU, the number of hidden units used is 50. Finally, fastText is employed as fastText has given better performance in recent day applications. The value of parameters for fastText are learning rate of 0.1, dimension of 128, minimum word count of 1, 100 epochs, and 2 N-grams. The average performance metrics of all the models for binary and multiclass data set are reported in Table \ref{Tab:5}. Among all, CNN-GRU along with Keras word embedding has performed better in both binary and multiclass classification. For all the models, the training  data set is used for training the models and testing data set is used to test the trained models.

\section{Conclusion and Future Work}

Twitter is one of the most popular social networks, where users share their opinions on various topics. The tweet could be related to security. This work evaluates the performance of various text representation techniques along with various deep learning models for cyber threat indicators in the Twitter stream. CNN-GRU with Keras embedding performed better than any other architecture in both binary as well as multiclass classification. The best part about the proposed architecture is that it does not require any feature engineering technique to be employed. Present and future work focus on event tracking and event detection of cyber threats using social media resources like Twitter, Facebook, etc.

\section*{Acknowledgements}

This research was supported in part by Paramount Computer Systems and Lakhshya Cyber Security Labs. We are grateful to NVIDIA India, for the GPU hardware support to research grant. We are also grateful to Computational Engineering and Networking (CEN) department for encouraging the research.

\end{document}